\documentclass[times, twocolumn,final,authoryear]{elsarticle}

\usepackage{prletters}
\usepackage{graphicx,subfigure}
\usepackage[labelfont=bf]{caption}
\usepackage{lipsum}
\usepackage{natbib}

\setcitestyle{numbers,square}

\journal{Pattern Recognition Letters}

\begin{document}

\begin{frontmatter}

\title{GaitPrivacyON: Privacy-Preserving Mobile Gait Biometrics using Unsupervised Learning\tnoteref{t1,t2}}
%\tnotetext[t1]{This work has been supported by project PriMa (MSCAITN-2019-860315).}
%\tnotetext[t2]{The second title footnote which is a longer
%text matter to fill through the whole text width and
%overflow into another line in the footnotes area of the
%first page.}
\author[1,2]{Paula Delgado-Santos}
%\fnref{fn1}}%
\cortext[cor3]{Corresponding author.}
\ead{p.delgado-de-santos@kent.ac.uk}
\author[2]{Ruben Tolosana}
\ead{ruben.tolosana@uam.es}
\author[1]{Richard Guest}
\ead{r.m.guest@kent.ac.uk}
\author[2]{Ruben Vera-Rodriguez}
\ead{ruben.vera@uam.es}
\author[1]{Farzin Deravi}
\ead{f.deravi@kent.ac.uk}
\author[2]{Aythami Morales}
\ead{aythami.morales@uam.es}
%\author[2]{CV Radhakrishnan\fnref{fn2}}
%\ead{cvr@sayahna.org}
%\author[3]{CV Rajagopal\fnref{fn1,fn3}}
%\ead[url]{www.stmdocs.in}

%\cortext[cor3]{Corresponding Author.}
%\cortext[cor2]{}
%\fntext[fn1]{E-mail address:}
%\fntext[fn2]{Another author footnote, this is a very long
%footnote and it should be a really long footnote. But this
%footnote is not yet sufficiently long enough to make two
%lines of footnote text.}
%\fntext[fn3]{Yet another author footnote.}
\address[1]{School of Engineering, University of Kent}
\address[2]{Biometrics and Data Pattern Analytics Lab, Universidad Autonoma de Madrid}
%\address[3]{STM Document Engineering Pvt}

\begin{abstract}

%This study proposes PrivSenseNet, a novel biometric mobile gait recognition approach that provides good performance while being able to remove sensitive information from the data. It comprise two modules: \textit{i)} TransforNet, an Autoencoder (AE) which transforms the raw data into a new format; and \textit{ii)} a subject verification system which is based on Convolutional Neural Networks (CNNs) and Long-Short Term Memory Recurrent Neural Networks (LSTM-RNNs) with a Siamese architecture. The main advantage of the proposed system is that apart from maintaining high the biometric gait verification task, it also focuses on the protection of sensitive data such as gender or the activity being carried out by the subject. This property of PrivSenseNet solves some of the drawbacks that have appeared in the state of the art. Numerous studies have shown that biometric gait verification systems perform well but, at the same time, it is possible to extract sensitive information from the subject. Also, we present remarkable experimental results for the biometric gait verification task while the recognition task of sensitive data remains practically random, showing the high potential of PrivSenseNet. To the best of our knowledge, it is the first biometric gait verification system that addresses privacy concerns for sensitive data in which training is performed by unsupervised learning. Therefore, all sensitive attributes are hiding in a single transformation.

Numerous studies in the literature have already shown the potential of biometrics on mobile devices for authentication purposes. However, it has been shown that, the learning processes associated to biometric systems might expose sensitive personal information about the subjects. This study proposes GaitPrivacyON, a novel mobile gait biometrics verification approach that provides accurate authentication results while preserving the sensitive information of the subject. It comprises two modules: \textit{i)} two convolutional Autoencoders with shared weights that transform attributes of the biometric raw data, such as the gender or the activity being performed, into a new privacy-preserving representation; and \textit{ii)} a mobile gait verification system based on the combination of Convolutional Neural Networks (CNNs) and Recurrent Neural Networks (RNNs) with a Siamese architecture. The main advantage of GaitPrivacyON is that the first module (convolutional Autoencoders) is trained in an unsupervised way, without specifying the sensitive attributes of the subject to protect. Two experimental studies have been examinated: \textit{i)} MotionSense and MobiAct databases; and \textit{ii)} OU-ISIR database. The experimental results achieved suggest the potential of GaitPrivacyON to significantly improve the privacy of the subject while keeping user authentication results higher than 96.6\% Area Under the Curve (AUC). To the best of our knowledge, this is the first mobile gait verification approach that considers privacy-preserving methods trained in an unsupervised way.

\end{abstract}


\end{frontmatter}

\section{Introduction}
\label{sec:introduction}

\begin{figure*}[htp]
\begin{center}
   \includegraphics[width=0.7\linewidth]{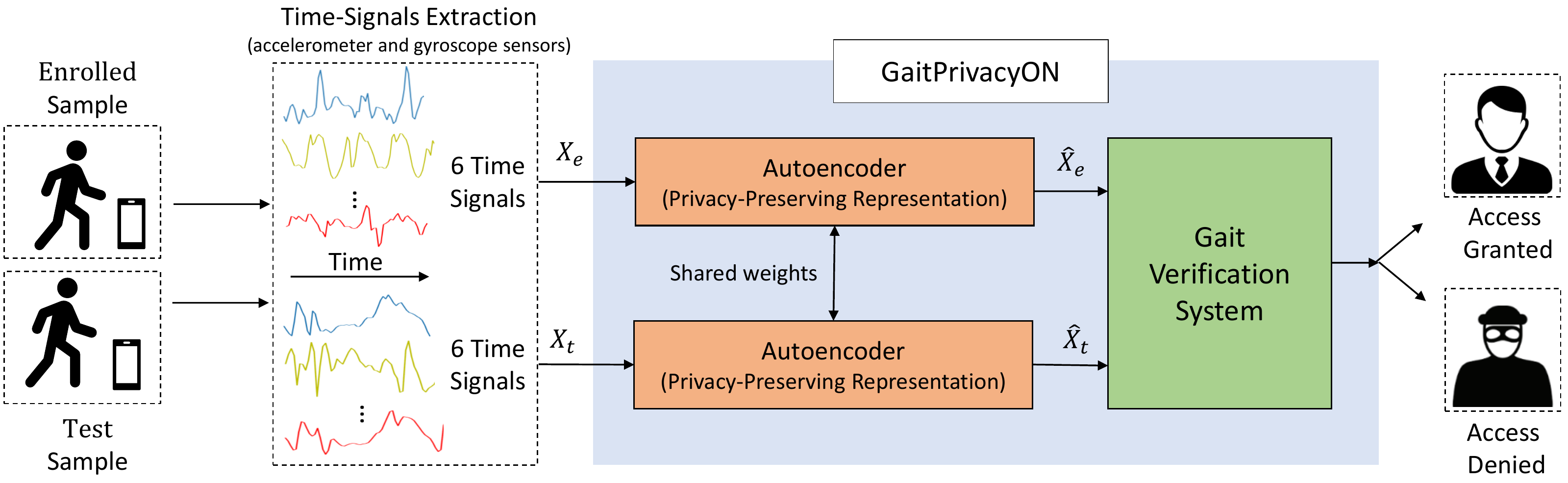}
\end{center}
   \caption{Diagram of GaitPrivacyON, which comprises two modules: \textit{i) } two Autoencoders that are in charge of removing automatically the sensitive data; and \textit{ii)} a gait verification system. Time signals extracted from the accelerometer and gyroscope sensors of the mobile devices are considered as input to GaitPrivacyON.  ${X}_{e}$: Enrolled sample, ${X}_{t}$: Test sample, $\widehat{X}_{e}$: Transformed enrolled sample, $\widehat{X}_{t}$: Transformed test sample.}
\label{fig:GaitPrivacyON}
\end{figure*}

The use of biometrics on mobile devices is currently one of the most popular authentication approaches \cite{boakes2019exploring, maiorana2011keystroke}. In particular, behavioural biometrics, which are based on the way subjects perform actions such as writing \cite{de2021air} and walking \cite{PAMI_RVera2018}, allow the recognition in a passive way through smart devices, for example, using the accelerometer and gyroscope data \cite{ellavarason2020touch,acien2019multilock}.

Despite the popularity of mobile behavioural biometrics, the data acquired can contain a large amount of personal and sensitive information such as demographics (e.g., gender, age, ethnicity, etc.)  or the activity the subject is performing (e.g., walking, sitting, etc.) \cite{2021_ArXiv_ChildCIdb_Tolosana}. As a result, this technology might be considered as an invasion of personal privacy. 
%miguel2016predicting
%piciucco2021biometric

Privacy is a concept that has been defined in numerous ways \cite{delgado2021survey}, one example of which is the recent General Data Protection Regulation (GDPR) of the European Union \cite{GDPR}. This defines personal data as ``any information relating to an identified or identifiable natural person". Within this set of data, there is a subgroup called sensitive data which includes ``racial or ethnic origin, political opinions, religious or philosophical beliefs, trade union membership, genetic data, biometric data for the purpose of uniquely identifying a natural person, health data or data concerning the individual's sex life or sexual orientation". The automatic processing of such data without the explicit consent of the subject for any specific purpose is prohibited. \hfill \break
\\
\\

%In the field of biometrics there are different studies that focus on hiding this sensitive data. Most of them focus on hiding only one kind of sensitive information. In contrast, this work privatises several types. In turn, state-of-the-art studies use adversarial learning for this purpose \cite{garofalo2019data, morales2020sensitivenets}. These works need to use information from sensitive data to train their models, introducing a new risk of leakage. In order to avoid this possibility, we propose a different model, based on the work of \cite{zhang2021preventing}. This model applies data transformation and noise addition without the need for sensitive data in the system.

The main contributions of this study are:

\begin{itemize}
    \item A novel mobile gait biometrics verification approach, GaitPrivacyON, that provides accurate authentication results while preserving the privacy of the subject. Fig. \ref{fig:GaitPrivacyON} represents the general diagram of our proposed approach. It comprises two modules: \textit{i)} two convolutional Autoencoders with shared weights that transform the biometric raw data into a new privacy-preserving representation (e.g., gender or activity), and \textit{ii)} a mobile gait verification system based on a combination of Convolutional Neural Networks (CNNs) and Recurrent Neural Networks (RNNs) with a Siamese architecture.
    %system that uses this new data representation in the mobile gait biometrics verification task
    \item An in-depth quantitative analysis of GaitPrivacyON over three popular databases in the field of gait recognition, MotionSense \cite{malekzadeh2018protecting}, MobiAct \cite{vavoulas2016mobiact}, and OU-ISIR \cite{Ngo2019OUISIR}, achieving accurate verification results (higher than 96.6\% Area Under the Curve, AUC) while reducing the recognition rate of sensitive data to $\sim$50\% AUC.
    %\item Two novel machine learning systems able to predict with high accuracy (99\% AUC) the gender and activity of the subject while using the mobile device.
    \item To the best of our knowledge,  this is the first mobile gait verification approach that considers privacy-preserving methods trained in an unsupervised way.
\end{itemize}    

The remainder of the paper is organised as follows. Sec. \ref{sec:RelatedWork} summarises previous studies in the field. Sec. \ref{sec:proposedmethod} explains all details of our proposed GaitPrivacyON approach. Sec. \ref{sec:databases} summarises the databases considered. Sec. \ref{sec:experimentalprotocol} and \ref{sec:experimentalresults} describes the proposed experimental protocol and results, respectively. Finally, Sec. \ref{sec:conclusions} draws the final conclusions.

\section{Related Work}
\label{sec:RelatedWork}

\subsection{Mobile Gait Biometrics}

Gait biometric recognition allows individuals to be authenticated based on the way they walk. It is a unique characteristic among individuals due to the specific arm swing amplitude, step frequency and length \cite{liu2021recent}. This characteristic can be easily detected in several ways. One of them is from Inertial Measurement Units (IMU), e.g., accelerometer and gyroscope \cite{liu2021recent}, which enables gait biometrics authentication from mobile devices. An example of this was presented by Mantyjarvi \textit{et al.} in \cite{mantyjarvi2005identifying}. Gait biometrics data captured by the accelerometer was used in a template matching and cross-correlation framework, achieving together, 7\% of Equal Error Rate (EER). Many researchers followed this method, proposing new studies in the literature as described in the review of Sprager and Juric \cite{sprager2015inertial}.

In recent years, Deep Learning (DL) approaches have dominated the field of gait recognition, being possible to extract more discriminative and robust features.  Gadaleta and Rossi created in \cite{gadaleta2018idnet} one of the first systems based on DL using CNNs. The authors used universal feature extractors for gait biometrics recognition with misclassification rates of $<$ 0.15\%. Their results showed that CNN-based systems learn more useful statistical features, achieving better performance than previous methods with pre-defined and often arbitrary features. 

In addition, RNNs is one of the most powerful DL techniques for temporal sequences \cite{2021_TBIOM_DeepSign_Tolosana, 2021_AAAI_DeepWriteSYN}. Ackerson \textit{et al.} proposed a new approach in which the OU-ISIR dataset was used \cite{ackerson2021applications}. The authors developed one of the first approaches to use a type of RNN, Long Short-Term Memory (LSTM), achieving an EER of 7.55\%.
%Fernandez-Lopez \textit{et al.} analysed accelerometer and gyroscope data to perform a subject recognition model \cite{fernandez2019recurrent}. A dataset with 774 subjects was used. The algorithm processed the signals by extracting the gait cycles and inputting them into an RNN, achieving an EER of 11.48\%. However, the data used came from a sensor on a belt attached to the waist. That work did not present a natural situation of holding a phone. In order to have a more realistic scenario, Ackerson \textit{et al.} proposed a new approach in which the OU-ISIR dataset was used \cite{ackerson2021applications}. This dataset comprises accelerometer and gyroscope data acquired from IMU sensors and accelerometer data from a smartphone. The authors developed one of the first approaches to use a type of RNN, Long Short-Term Memory (LSTM). The authors achieved an EER of 7.55\%. Watanabe and Kimura collected mobile accelerometer data from 21 individuals in a more common scenario \cite{watanabe2020gait} . The subjects were carrying the device in their pocket or hand while walking. Following the DL advances, the authors proposed a LSTM framework, achieving an accuracy of 97\%. 

%More approaches have been recently proposed in this sense considering RNN architectures \cite{zaroug2021prediction, tran2021multi}.

Another interesting approach was proposed by Zou \textit{et al.} in \cite{zou2020deep}. An hybrid DL model combining CNNs and LSTM for more robust features was created. The proposed model brought together the advances of CNNs (extracting convolutional maps with more discriminative features) and RNNs (processing features as temporal sequences). Mobile devices in the wild were considered, with 118 subjects and data extracted from the accelerometer and gyroscope, obtaining an accuracy of 93.7\%.

\subsection{Privacy-Preserving Methods}

Privacy-preserving concerns are becoming increasingly important nowadays due to the new privacy laws and regulations. Therefore, many researchers have extensively studied the field in the last decade \cite{delgado2021survey}. 

%(e.g., downstairs,  upstairs,  jogging,  and  walking) 
%Osia \textit{et al}. focused on activity recognition \cite{osia2020hybrid}. The authors developed a Siamese CNN split between an IoT device, holding only the necessary information, and the cloud. The authors evaluated the model with the information exposed to the cloud service, managing to increase the EER in the gender classification task by 14\%.

In the human-activity recognition field, Iwasawa \textit{et al}. proposed a model with an adversarial subject classifier and a regular activity-classifier based on CNNs \cite{iwasawa2017privacy}. The authors managed to privatise the subject's discriminative information by 40\% while keeping accurate activity recognition performance. Malekzadeh \textit{et al}. presented in \cite{malekzadeh2018protecting} a feature learning architecture that provides privacy-preserving data transmission and a new dataset for activity and attribute recognition collected from motion sensors. Their system was based on Generative Adversarial Networks (GANs), achieving a 45.8\% reduction in accuracy in the gender classification task while the activity recognition task only decreased by 1.37\%. Zhang \textit{et al.} proposed a new framework for activity recognition and privacy-preserving of sensitive data \cite{zhang2021preventing}. The authors wanted to avoid the need for massive collection of sensitive data for model training. For this purpose, an unsupervised learning training for the privacy-preserving task was performed. The framework was treated by a transformation of the data together with a noise addition consisting of an Autoencoder and a CNN. Results of 56.79\% accuracy was achieved for gender classification while the activity recognition task remained almost untouched.

In the gait biometrics verification field, Garofalo \textit{et al}. developed a Siamese CNN framework \cite{garofalo2019data}. The authors decreased the F1-score in the gender recognition task from 73\% to 52\% while loosing from 90.93\% to 85.28\% of accuracy in the gait verification task. An adversarial learning technique was used.

Several techniques have also been applied to the image field but at the feature representation level. Therefore, these approaches could also be adapted for time signals. Terh{\"o}rst \textit{et al.} proposed an unsupervised approach based on similarity-sensitive noise transformations \cite{terhorst2019unsupervised}. That approach added noise to the feature representations. Experiments showed how attackers with prior knowledge about the privacy mechanism (added cosine noise) decreased the accuracy of gender estimation performance with logistic regression $\sim$17\%. Identity recognition performance only increased by $\sim$5\% EER. In \cite{terhorst2019suppressing} Incremental Variable Elimination (IVE) algorithm was proposed. The model was used to suppress binary and categorical attributes in biometric templates. The model, through decision tree training, managed to decrease the gender Correct Overall Classification Rate (COCR) by 20\% but only increasing the EER in the identity recognition task by 1.4\%. %Instead of masking sensitive attributes, the authors in \cite{bortolato2020learning} presented an approach that disentangles feature representations so sensitive information can be removed. That paper presented Privacy-Enhancing Face-Representation learning Network (PFRNet), an Autoencoder that achieved a latent representation in which increasing the EER in the identity recognition task by 2.7\%, increases the Fraction of Incorrectly Classified images (FIC) in the gender recognition task to 43.5\%.
Morales \textit{et al.} created SensitiveNets \cite{morales2020sensitivenets}. Its main purpose was to set aside sensitive information in decision making in order to ensure fairness and transparency. It was tested on feature representations of face images by defining and minimising its own loss function. SensitiveNets achieved representations that reduced the gender and ethnicity classification tasks to 54.6\% and 53.5\% respectively, decreasing recognition task accuracy only 2.6\%.

The previous gait verification approaches presented in the literature can preserve specific sensitive attributes \cite{garofalo2019data}, but they require a large volume of labelled data for training. On the contrary, GaitPrivacyON considers unsupervised learning for the privacy preserving of the subjects without specifying the sensitive attributes to protect. Therefore, this avoids any model inference and provides greater protection than the models presented in the literature. Also, this approach allows the hiding of all sensitive attributes using a single transformation.

\section{Proposed Approach: GaitPrivacyON}
\label{sec:proposedmethod}

%This study proposes a novel framework in order to prevent leakage of sensitive information in a user verification model. For this purpose, a noise addition and a data transformation systems are included. This scheme has been verified with temporal sequences extracted from the motion sensors of a mobile device (eg. accelerometer and gyroscope), but it could be applied to any other type of temporal sequence, such as keystroke. In this section, three new models are proposed: \textit{i)} a novel user verification system based on Convolutional Neural Networks (CNN) and RNN (Recurrent Neural Networks), \textit{ii)} a novel sensitive attribute inference system based on CNN, and \textit{iii)} PrivSenseNet, a novel privacy prevention system for sensitive attributes based on an autoencoder (AE) and the system proposed in \textit{i)}.

%In order to overcome these limitations, the approach presented here has a Siamese architecture with two inputs. Also, is based in a gait biometrics verification scenario instead of activity recognition.

\begin{figure*}[t!]
\begin{center}
   \includegraphics[width=0.8\linewidth]{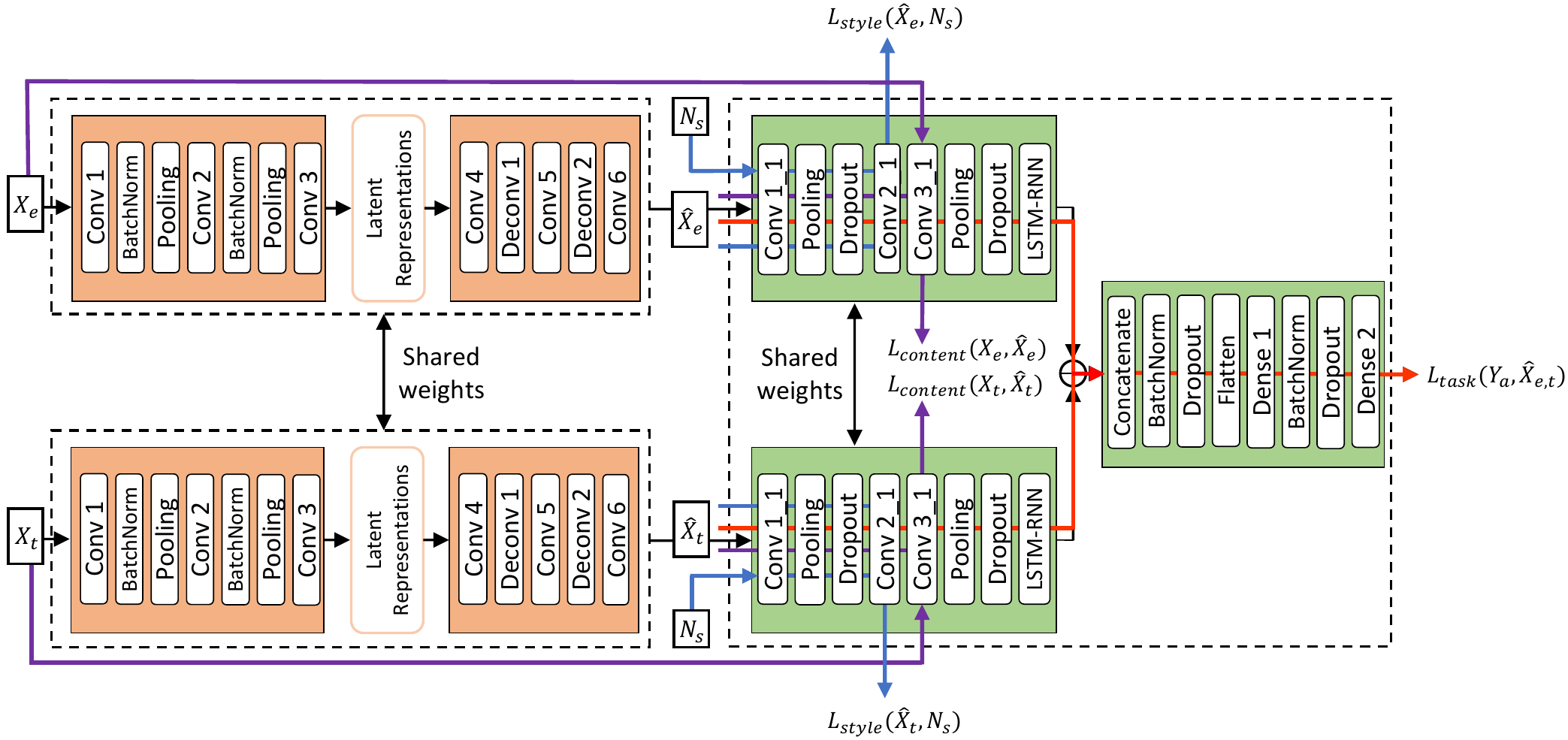}
\end{center}
   \caption{Architecture and training losses ($\mathcal{L}_{content}$, $\mathcal{L}_{style}$, $\mathcal{L}_{task}$) considered in GaitPrivacyON. ${X_e}, {X_t}$: Raw time signals; $\widehat{X_e}, \widehat{X_t}$: Transformed time signals; $N_s$: Random noise; $Y_{a}$: label of the gait verification task.}
\label{fig:GaitPriv_losses}
\end{figure*}

Fig. \ref{fig:GaitPrivacyON} shows the general diagram of the proposed privacy-preserving approach. Six time signals are originally acquired from the mobile device as raw data comprising the three axes of the accelerometer and gyroscope. GaitPrivacyON considers a Siamese architecture that is used to learn the similarity between two different biometric templates from the same (genuine) or different (impostor) subject \cite{maiorana2019eeg}. GaitPrivacyON comprises two modules: \textit{i)} two convolutional Autoencoders with shared weights that transform the biometric raw data into a new privacy-preserving representation (Sec. \ref{sec:Autoencoders}); and \textit{ii)} a mobile gait verification system based on the combination of CNNs and RNNs with a Siamese architecture (Sec. \ref{sec:GaitVerifSystem}). For the training, we adapted the key aspects presented in the image style transformation field \cite{zhang2021preventing}. The details are explained in Sec. \ref{sec:training}. GaitPrivacyON is an improved adaptation of the approach presented in \cite{zhang2021preventing}. We clarify next the main changes: 
\begin{itemize}
    \item GaitPrivacyON is based on gait biometric verification while the approach presented in  \cite{zhang2021preventing} is based on activity recognition. As a result, our approach focuses on verification (1:1) rather than identification (1:N).
    \item Regarding the Autoencoders considered in GaitPrivacyON (see details in Sec. \ref{sec:Autoencoders}), while TransNet only has a single Autoencoder, two Autoencoders are considered in GaitPrivacyON, sharing their weights through a Siamese architecture. In addition, in order to extract more discriminative features and improve the training, we have considered batch normalization and increased the complexity of the network using more convolutional layers.
    \item The Gait Verification System proposed in GaitPrivacyON (see details in Sec. \ref{sec:GaitVerifSystem}) has several differences compared with LossNet  \cite{zhang2021preventing}. LossNet is based only on convolutional layers. The system presented in this work considers both convolutional and recurrent layers following the state of the art in gait biometrics \cite{zou2020deep}. This improves the performance of the system and makes our system more robust.
    
\end{itemize}

\subsection{Autoencoders}\label{sec:Autoencoders}

Fig. \ref{fig:GaitPriv_losses} (orange colour) provides a graphical representation of the proposed module. It comprises two convolutional Autoencoders with the same architecture and shared weights. The inputs of each are: enrolled sample ($X_e$) and test sample ($X_t$), and the outputs:  transformed enrolled sample ($\widehat{X_e}$) and transformed test sample ($\widehat{X_t}$), respectively. The architecture of both is composed of a sequence of 1 $\times$ 3 convolutional filters, coupled with ReLU activation functions. In the encoder, after each convolutional layer, batch normalization and 1 $\times$ 2 max-pooling layers are used to decrease the size of the activation map. In the decoder, after each convolutional layer, a deconvolutional layer is used with 1 $\times$ 3 strides of the convolution. The activation function of the last convolutional layer is linear. In GaitPrivacyON, the loss of the main task ($\mathcal{L}_{task}$) is in charge of training the Autoencoders to extract useful transformed data ($\widehat{X}$). This loss is considered together with the loss of content ($\mathcal{L}_{content}$), responsible for retaining authentication information, and the loss of style ($\mathcal{L}_{style}$), which removes sensitive data by introducing uniform random noise ($N_s$).
 
\subsection{Gait Verification System}\label{sec:GaitVerifSystem}

Fig. \ref{fig:GaitPriv_losses} (green colour) provides a graphical representation of the architecture proposed for gait verification ($\varphi$). In particular, we have adapted the approach originally presented by Zou \textit{et al.} in \cite{zou2020deep} to our specific case (privacy-preserving gait verification). 
It is based on a novel Siamese architecture with two inputs: transformed enrolled sample ($\widehat{X_e}$) and transformed test sample ($\widehat{X_t}$). The inputs are reshaped including one new dimension. Unlike the method proposed in \cite{zou2020deep}, the architecture is composed of a sequence of 1 $\times$ 3 two-dimensional convolutional filters, coupled with ReLU activation functions. After 3 convolutional layers, batch normalization, 1 $\times$ 2 max-pooling, and dropout with a probability of 0.5 are used. A reshaping layer is included to return to the shape of the time domain signals following by a bi-directional LSTM layer with 50 units. The dense layer has a size of 400 with a sigmoid activation function.

%The proposed gait verification system is first trained end-to-end and fixed with pre-trained weight parameters from the biometric raw data.

%Finally, while the work shown by Zou only learns features while the subject is walking, our system is trained with several activities (downstairs, upstairs, jogging and walking), so it has to learn more complex features.

\subsection{Traning}\label{sec:training}

GaitPrivacyON is trained following the idea proposed in the image style transformation field \cite{johnson2016perceptual}. One image can be divided into two parts: \textit{i)} the \textit{content}, i.e., what is in the image, and \textit{ii)} the \textit{style}, i.e., how the image is illustrated. In our particular application of gait biometrics verification, the content is the unique information that allows to verify the identity of the subject whereas the style is the sensitive information of the subject that can be considered for other purposes not related to the authentication. This sensitive information may include the person's gender, age, ethnicity, or the activity the subject is performing while using mobile devices \cite{iwasawa2017privacy}.

Following this idea, three different loss functions have been considered from the work presented in \cite{zhang2021preventing}: \textit{task loss} ($\mathcal{L}_{task}$), \textit{content loss} ($\mathcal{L}_{content}$), and \textit{style loss} ($\mathcal{L}_{style}$).

The \textit{task loss} ($\mathcal{L}_{task}$) helps the system to maintain its usefulness in the main task of gait verification. We consider a categorical cross-entropy that compares the transformed data ($\widehat{X}$) with the biometric raw data ($X$). The \textit{task loss} can be defined as:
\begin{equation}
    \mathcal{L}_{task} (Y_{a},\widehat{X})= -Y_{a}log(\varphi(\widehat{X}))
\end{equation}

\noindent where $Y_{a}$ and $\varphi(\widehat{X})$ are the label and the predicted probability of the gait verification task, respectively. 

The \textit{content loss} ($\mathcal{L}_{content}$) measures the content (i.e., the authentication information) that the transformed data ($\widehat{X}$) and the biometric raw data ($X$) have in common. For this aim, we use the Euclidean distance to compare the feature maps provided by the \textit{i}-layer of the $\varphi$ network when using both the biometric raw data and the transformed data as input. In our case, we use the feature maps obtained behind \textit{Conv3\_1} layer in Fig. \ref{fig:GaitPriv_losses}. This was decided experimentally. The \textit{content loss} is defined as:
\begin{equation}
    \mathcal{L}_{content}^{i}(X,\widehat{X}) = \frac{1}{C_{i} H_{i}
    W_{i}}\left \| \varphi _{i}(\widehat{X}) -\varphi _{i}({X}) \right \|_{2}^{2}
\end{equation}

\noindent where \textit{i} is the layer and $C_{i} \times H_{i} \times W_{i}$ is the shape of the feature map obtained after this layer. Comparing feature maps ensures that the content of the biometric raw data and the transformed data are similar but do not have to be identical.

The \textit{style loss} ($\mathcal{L}_{style}$) is responsible for maintaining the transformed data ($\widehat{X}$) unstyled, thus avoiding the extraction of any sensitive information automatically. For this purpose, we want to modify the style of the data by uniform random noise ($N_s$) with range [-20, 20] as done by \cite{zhang2021preventing}. We consider the Gram matrix ($G$) to measure the style differences between feature representations. Random noise is introduced as the new domain, avoiding using any information from the sensitive data for its protection, creating an unsupervised learning framework. For this aim, both the transformed data and the random noise are fed into the trained gait verification system with the weights frozen. After that, the Gram Matrices of the feature maps obtained as output of the $i$-layer are compared.
The Gram Matrix can be defined as:
\begin{equation}
    G_i(X)_{c,c'} = \frac{1}{C_{i} H_{i}
    W_{i}} \sum_{h = 1}^{H_i}\sum_{w = 1}^{W_i}\varphi _{i}(X)_{h,w,c}\varphi _{i}(X)_{h,w,c'}
\end{equation}

\noindent where the shape of $\varphi_i(X)$ is $C_{i} \times H_{i} \times W_{i}$ and the shape of its Gran matrix ($G_i^{\varphi}$) is $\left | C_{i} \right | \times \left | C_{i} \right |$. $\varphi_i(X)$ can be interpreted as $C_{i}$ dimensional features for each $H_{i} \times W_{i}$ point, where $c$ and $c'$ are two different dimensions.

The \textit{style loss} measures the dissimilarity in style using the Frobenius squared norm of the difference of the Gram matrices of the transformed data ($\widehat{X}$) and the random noise ($N_s$). In our case, we have decided to use the feature maps obtained behind  \textit{Conv2\_1} in Fig. \ref{fig:GaitPriv_losses}. The \textit{style loss} can be defined as:
\begin{equation}
    \mathcal{L}_{style}^{i} (\widehat{X}, N_s) = \left \| G_i^{\varphi}(\widehat{X}) - G_i^{\varphi}(N_s) \right \|_{F}^{2}
\end{equation}
where $F$ denotes the Frobenius squared norm. By using deeper layers, the extracted features will be more similar.

The final loss function of GaitPrivacyON ($ \mathcal{L}_{total}$) would be a weighted sum of the losses $\mathcal{L}_{task}$, $\mathcal{L}_{content}$, and $\mathcal{L}_{style}$:
\begin{equation}
    \mathcal{L}_{total} = \alpha \mathcal{L}_{task} + \beta \mathcal{L}_{content} + \gamma \mathcal{L}_{style} 
\end{equation}
where $\alpha + \beta + \gamma = 1$.

\section{Databases}
\label{sec:databases}

\subsection{MotionSense Database}
The MotionSense database \cite{malekzadeh2018protecting} comprises accelerometer and gyroscope data collected with an iPhone 6s. A total of 24 subjects with information on gender, age, height, and weight, were obtained. The data was acquired while the subjects performed 4 different activities (walking up and down stairs, jogging, and walking). All the subjects had the mobile phone fixed in the front pocket of the trouser.

\subsection{MobiAct Database}

The MobiAct database \cite{vavoulas2016mobiact} comprises accelerometer, gyroscope and magnetometer data collected using a Samsung Galaxy S3. A total of 56 subjects performing the same 4 activities (walking up and down stairs, jogging, and walking) were captured. Data on gender, age, height, and weight of the subjects were acquired. Unlike the previous database, subjects had a free choice of placement of their device, simulating a realistic scenario.

\subsection{OU-ISIR Database}

The popular OU-ISIR database \cite{Ngo2019OUISIR} is considered in this study. It comprises accelerometer and gyroscope data collected from three inertial measurement units and a Motorola ME860 around the waist of the subject. In total, 744 subjects with gender and age data were captured. All subjects performed 4 activities (two flat walking, slope-up walking, and slope-down walking).

\section{Experimental Protocol}
\label{sec:experimentalprotocol}

%The main goal behind the experimental protocol design is to analyse and prove the potential of the GaitPrivacyON approach for gait biometrics scenarios.

GaitPrivacyON considers two main tasks: \textit{i)} gait biometrics verification, and \textit{ii)} privacy-preserving information, for which auxiliary machine learning systems must be implemented to detect the subject sensitive information, in our case, the gender and activity of the subject while using the mobile device. The specific details of the architecture are included in Sec. \ref{ssec:GenderActivitySystems}.

Regarding the training procedure, GaitPrivacyON first trains only the gait verification system using the biometric raw data ($X$) from the development dataset. For this first stage, binary cross-entropy is considered for the loss function. After that, we train our proposed GaitPrivacyON approach (only the Autoencoders module, the weights of the gait verification system are frozen) using the same development dataset. In this second stage, the total loss function ($L_{total}$) considered in GaitPrivacyON is a weighted sum of the losses $L_{task}$, $L_{content}$, and $L_{style}$, as described in Sec. \ref{sec:proposedmethod}. The specific details of the development and final evaluation datasets are provided in Sec. \ref{ssec:ProtocolMotionSenseMobAct} and Sec. \ref{ssec:ProtocolOUISIR}.

\begin{table}[tp!]
\centering
\caption{Architecture of the gender and activity inference systems. Prob- Probability. m- number of signals. SAC- Sensitive Attribute Classes.}
\resizebox{\columnwidth}{!}{
\begin{tabular}{lccccc}
\hline
\multicolumn{1}{c}{\textbf{Layer}} & \textbf{\begin{tabular}[c]{@{}c@{}}Input Size \\ ($H\times W\times F$)\end{tabular}} & \textbf{\begin{tabular}[c]{@{}c@{}}Kernel \\ ($H\times W$)\end{tabular}} & \textbf{Padding} & \textbf{Activation} & \textbf{Prob}  \\ \hline
Conv1$\_$1                           & m$\times$100$\times$1                     & 1$\times$3                   & Valid            & Relu                & -             \\
Conv1$\_$2                           & m$\times$98$\times$16                     & 1$\times$3                   & Valid            & Relu                & -             \\
Batch$\_$1                           & m$\times$96$\times$16                     & -                     & -                & -                   & -             \\
Pool$\_$1                            & m$\times$96$\times$16                     & 1$\times$2                   & Valid            & -                   & -             \\
Drop$\_$1                            & m$\times$48$\times$16                     & -                     & -                & -                   & 0.5           \\ \hline
Conv2$\_$1                           & m$\times$48$\times$16                     & 1$\times$5                   & Valid            & Relu                & -             \\
Batch$\_$2                           & m$\times$44$\times$32                     & -                     & -                & -                   & -             \\
Pool$\_$2                            & m$\times$22$\times$32                     & 1$\times$2                   & Valid            & -                   & -             \\
Drop$\_$2                            & m$\times$22$\times$32                     & -                     & -                & -                   & 0.5           \\ \hline
Dense$\_$1                           & \begin{tabular}[c]{@{}c@{}}m$\times$100\end{tabular}                 & -                     & -                & -                   & -             \\
Batch$\_$3                           & m$\times$100                       & -                     & -                & -                   & -             \\
Drop$\_$3                            & m$\times$100                       & -                     & -                & -                   & 0.5           \\ \hline
Dense$\_$2                           & m$\times$SAC                       & -                     & -                & -                   & -  
\\ \hline
\end{tabular}
}
\label{table:SensAttriNet}
\end{table}
 
\subsection{Gender and Activity Inference Systems} \label{ssec:GenderActivitySystems}

Table \ref{table:SensAttriNet} shows the architecture of the proposed gender and activity inference systems. Six time signals are originally acquired from the mobile device as raw data, the three axes of the accelerometer and gyroscope. The input data is in the same shape as in GaitPrivacyON. The architecture is composed of a sequence of 1 $\times$ 3 convolutional filters, coupled with ReLU activation functions. After some convolutional layers, batch normalization, 1 $\times$ 2 max-pooling, and dropout with a probability of 0.5 are used. The dense layer has a size of 100. For the gender recognition system, a sigmoid activation function is considered whereas softmax is considered for the activity recognition system. Finally, cross-entropy is used for the loss function.

\begin{figure}[t!]
\centering
\subfigure{\includegraphics[width=0.6\linewidth]{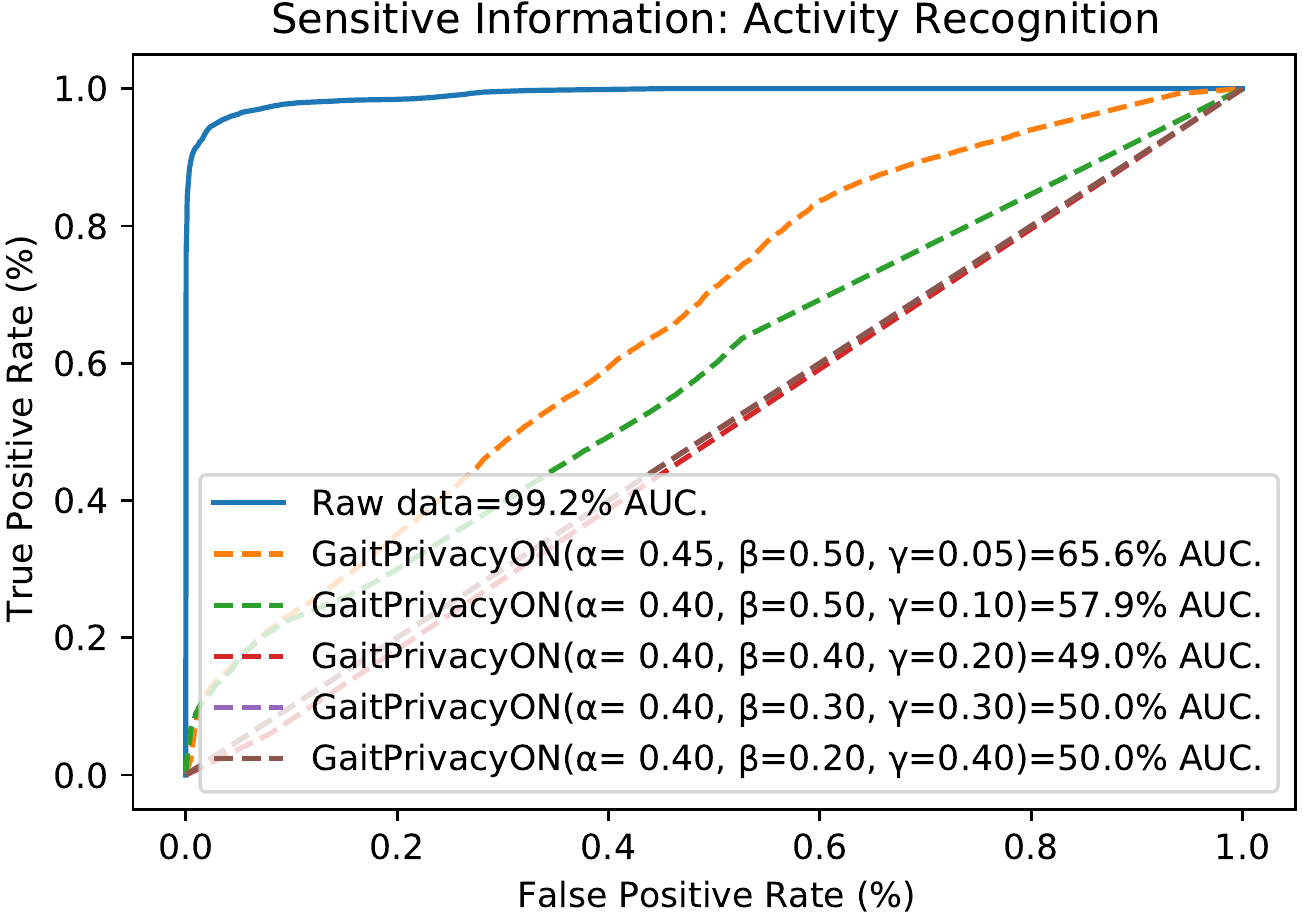}}
\subfigure{\includegraphics[width=0.6\linewidth]{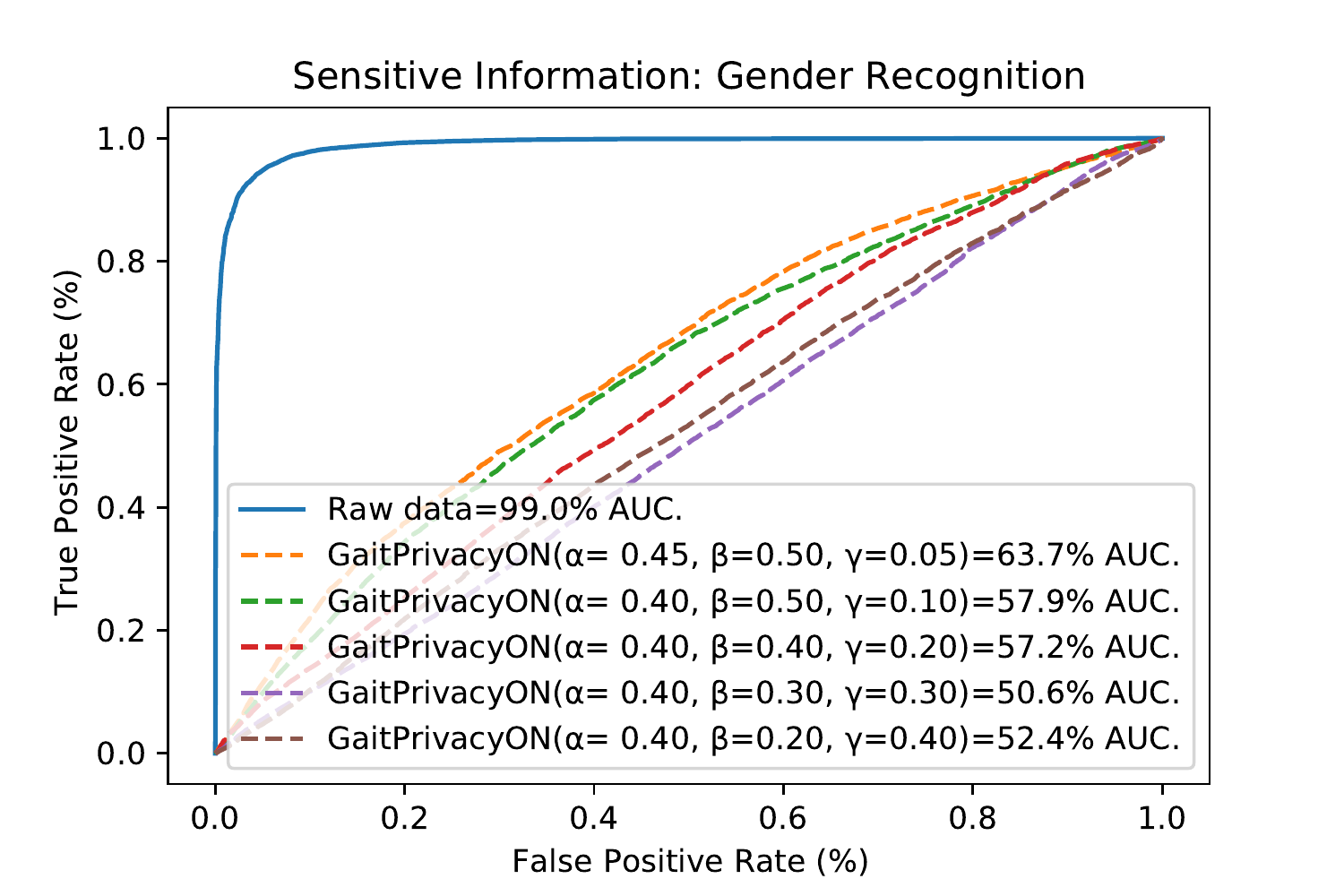}}
\subfigure{\includegraphics[width=0.6\linewidth]{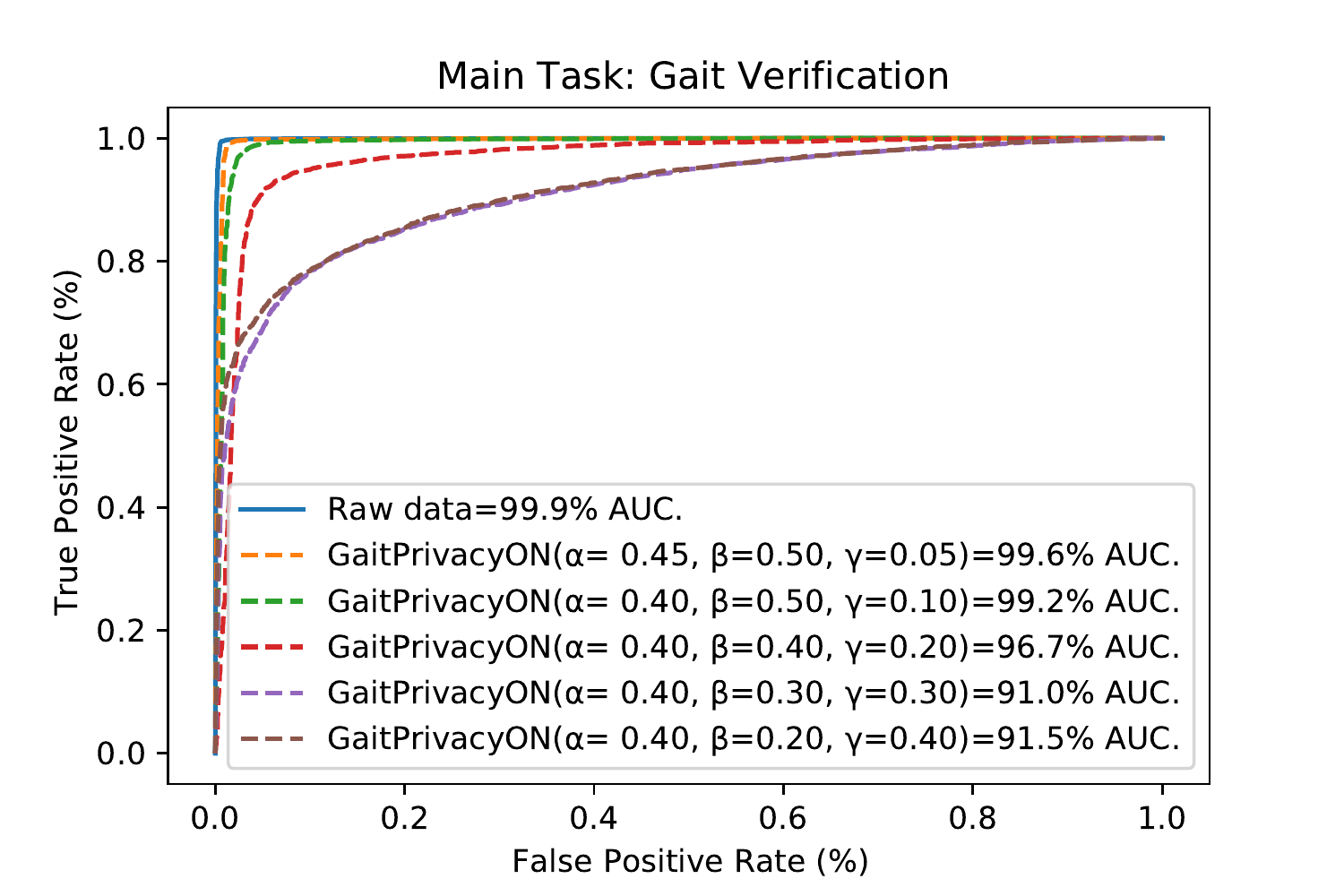}}
\caption{ROC curves and AUC (\%) results on the MotionSense and MobiAct evaluation dataset for the two scenarios considered: \textit{i)} Biometric raw data ($X$), and \textit{ii)} GaitPrivacyON ($\widehat{X}$). Different parameters ($\alpha, \beta, \gamma$) of GaitPrivacyON are tested in order to evaluate the results of the main task (gait verification) and the privacy-preserving information of the subject (activity and gender recognition).}
\label{fig:ROC_MotionMobi}
\end{figure}

\subsection{MotionSense \& MobiAct Databases} \label{ssec:ProtocolMotionSenseMobAct}
Our approach is trained with accelerometer and gyroscope time signals from both MotionSense and MobiAct databases. A total of 80 subjects (i.e., 24 from MotionSense and 56 from MobiAct) performing 4 different activities (walking up and down stairs, jogging, and walking) are considered in the experimental framework. The total database consists of 55 males and 25 females. In both databases the frequency sampling has been normalised to mean 0 and standard deviation 1, with a sampling frequency of 50 Hz. Each time signal comprises 100 samples. Also, we consider time windows of 2 seconds with an overlapping ratio of 75\%. The total database is divided into development and evaluation datasets, which contain different subjects with random selection. The development dataset, used for the training of GaitPrivacyON, has 70 subjects (85\% of the subjects have been used for training and the remaining part for validation). After training, the remaining 10 unseen subjects are used for the final evaluation. 
Regarding the gender and activity inference systems (see Sec. \ref{ssec:GenderActivitySystems}), we consider the same development and evaluation datasets described before, balancing the number of male and female subjects to avoid bias (5 males and 5 females in the final evaluation set). All subjects contain the same 4 activities.

\subsection{OU-ISIR Database} \label{ssec:ProtocolOUISIR}

GaitPrivacyON is trained with accelerometer and gyroscope time signals using the right-position inertial measurement unit, as it is more reliable according to \cite{Ngo2019OUISIR}. In the scenario of performing 4 different activities (two flat walking, slope-up walking, and slope-down walking), there are 492 subjects available (256 males and 236 females). The data have been normalised with mean 0 and standard deviation 1, with a sampling frequency of 100 Hz. Each time signal has a time window of 1 second, which is defined as 100 samples, and an overlapping between time windows of 75\%. This database is divided into development and evaluation, which comprises different subjects with random selection. For the training of GaitPrivacyON, the development dataset contains 80\% of the subjects (312 for training and 80 for validation). After the training, the remaining 20\% of the subjects (100 unseen subjects) are used for the final evaluation. 
Regarding the gender and activity inference systems (described in Sec. \ref{ssec:GenderActivitySystems}), we consider the same development and evaluation datasets described before, balancing the number of male and female subjects to avoid bias (50 males and 50 females in the final evaluation set). All subjects contain the same 4 activities.

\section{Experimental Results}
\label{sec:experimentalresults}

\subsection{Gender and Activity Inference from Biometric Raw Data}

In this first experiment we analyse the ability of machine learning systems to infer sensitive information of the user from the biometric raw data.

\subsubsection{MotionSense \& MobiAct Databases}

Fig. \ref{fig:ROC_MotionMobi} (top) shows the Receiver Operating Characteristic (ROC) curve together with the AUC of the activity recognition system (solid curve). The proposed system achieves 99.2\% AUC, differentiating the activity (walking up and down stairs, jogging, and walking) with precision.

Second, we analyse the results achieved by the proposed gender recognition system. The system has two clases: male and female. Fig. \ref{fig:ROC_MotionMobi} (middle) shows the ROC curve together with the AUC achieved by the gender recognition system (solid curve). As in the case of the activity task, the gender recognition system is able to differentiate the gender with 99.0\% AUC.

\subsubsection{OU-ISIR Database}

\begin{figure}[t!]
\centering
\subfigure{\includegraphics[width=0.6\linewidth]{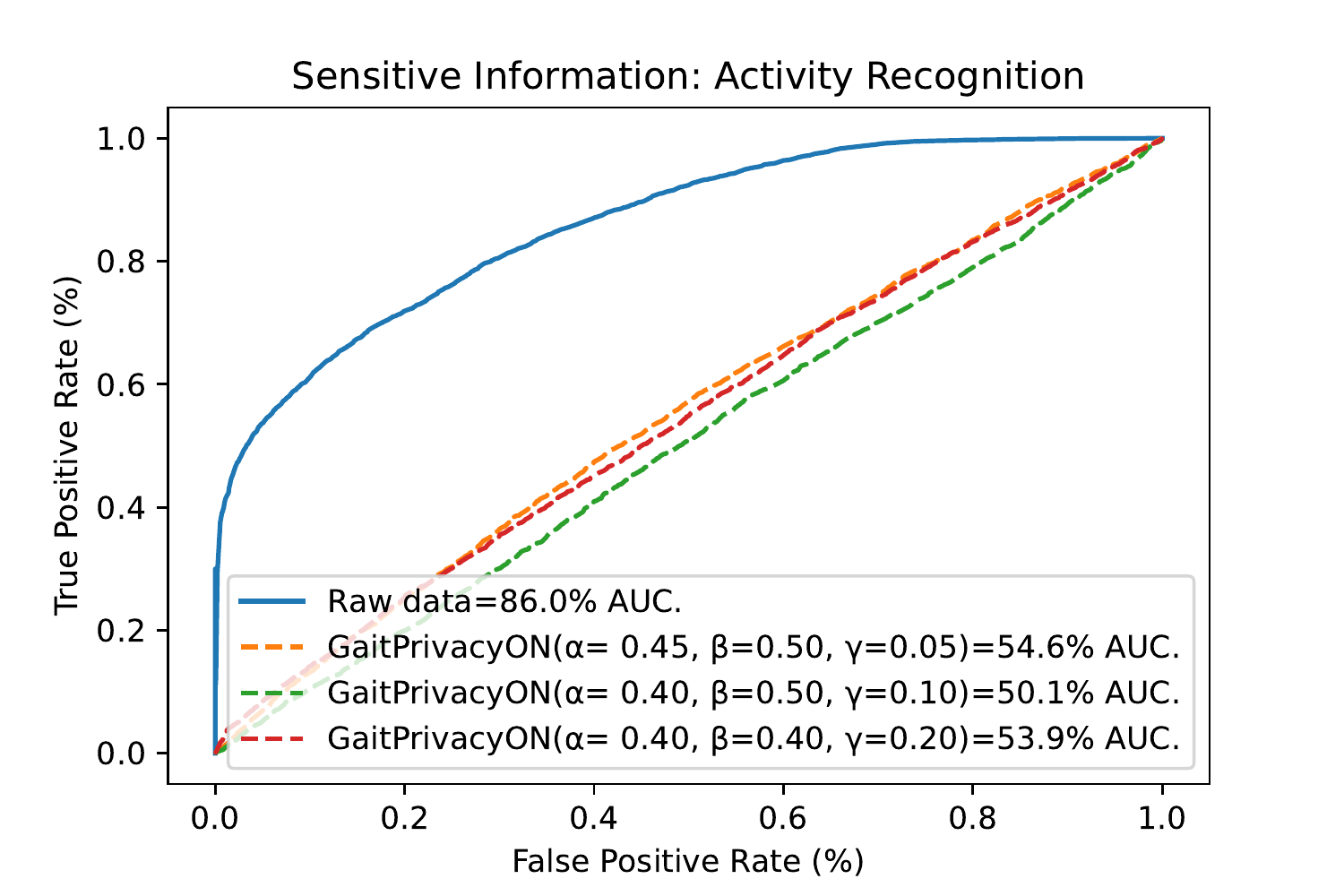}}
\subfigure{\includegraphics[width=0.6\linewidth]{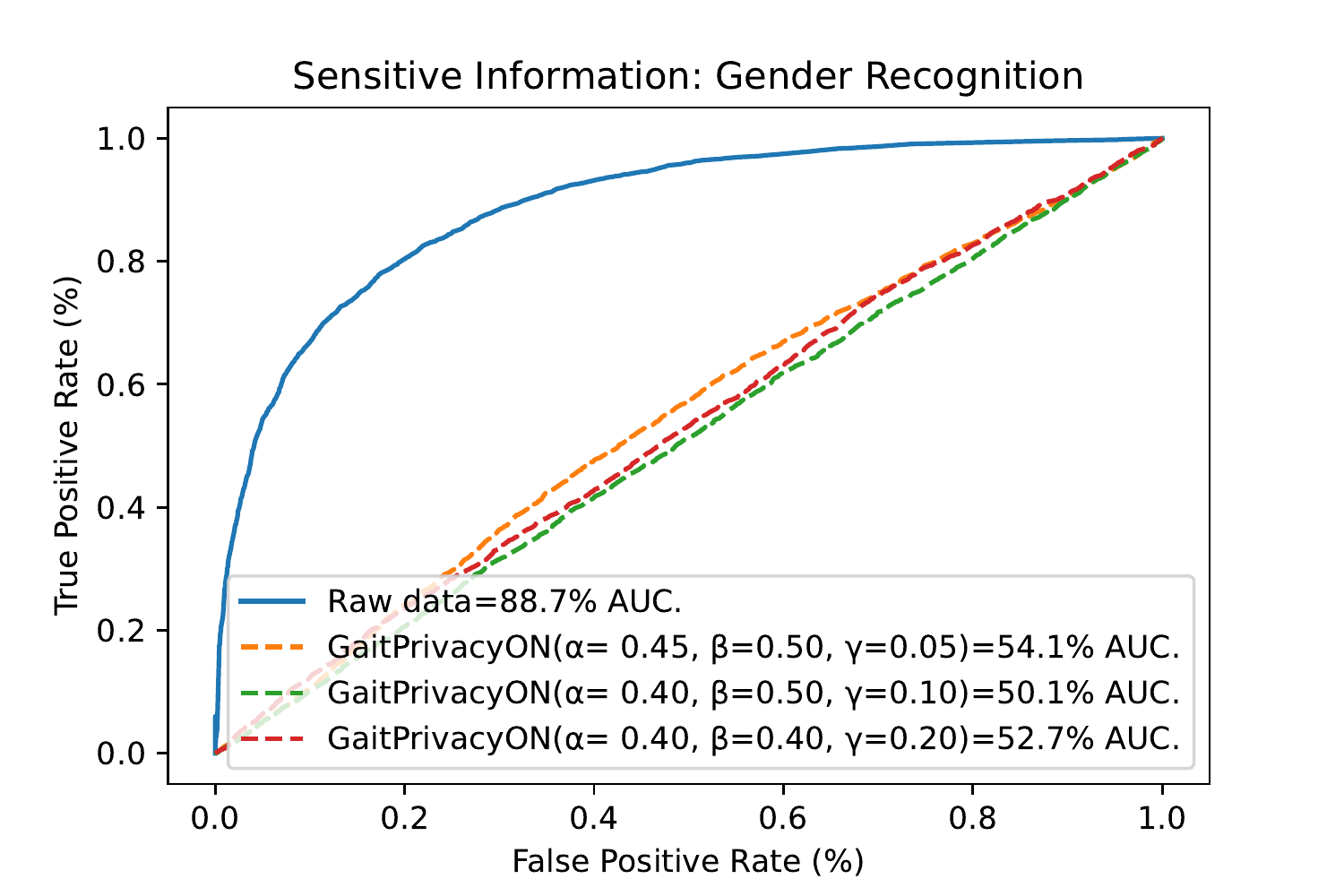}}
\subfigure{\includegraphics[width=0.6\linewidth]{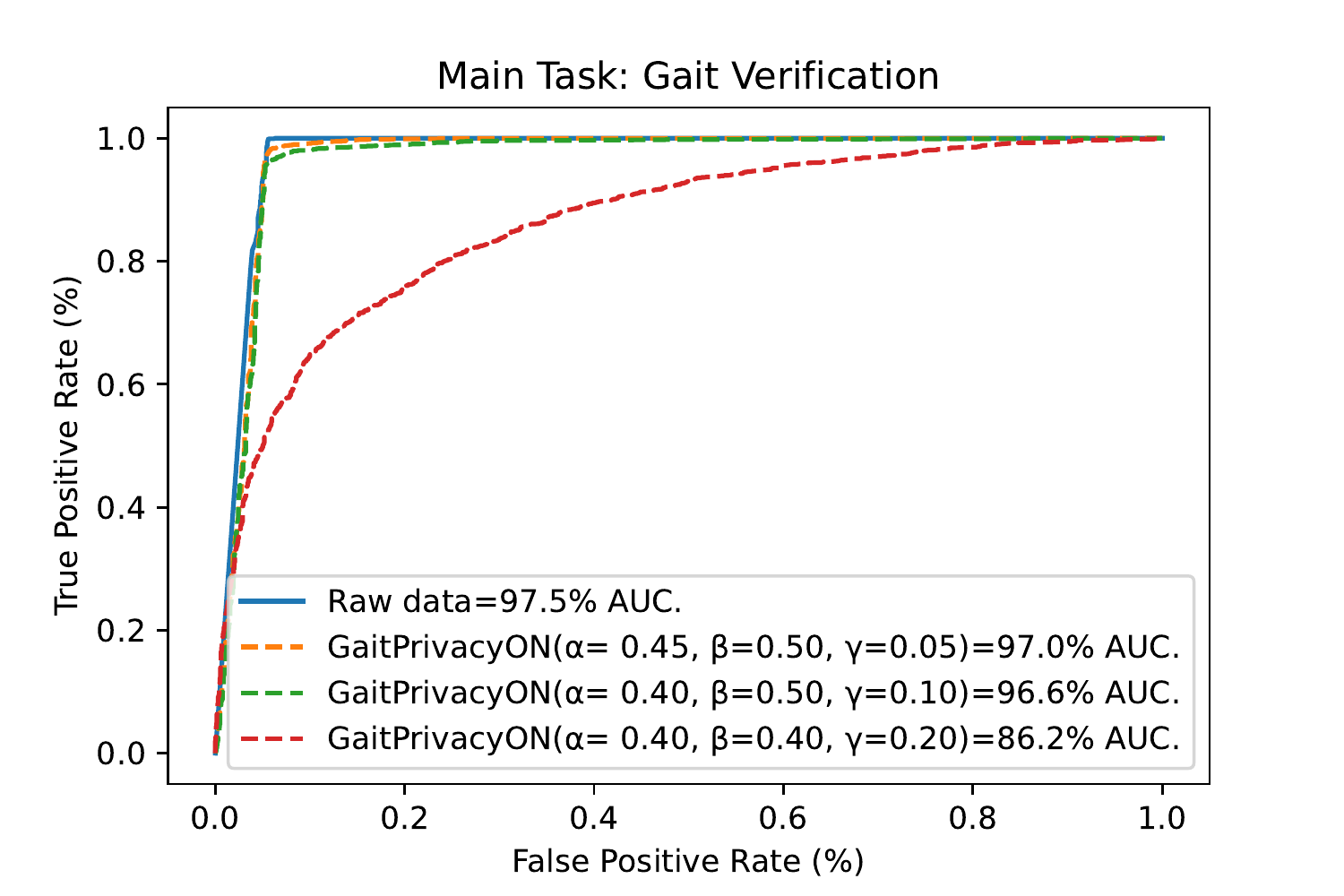}}
\caption{ROC curves and AUC (\%) results on the OU-ISIR evaluation dataset for the two scenarios considered: \textit{i)} Biometric raw data ($X$), and \textit{ii)} GaitPrivacyON ($\widehat{X}$). Different parameters ($\alpha, \beta, \gamma$) of GaitPrivacyON are tested in order to evaluate the results of the main task (gait verification) and the privacy-preserving information of the user (activity and gender recognition).}
\label{fig:ROC_OUISIR}
\end{figure}

Fig. \ref{fig:ROC_OUISIR} (top) shows the ROC curve together with the AUC result achieved by the activity recognition system (solid curve). Similar to the MotionSense and MobiAct databases, the system is able to achieve accurate results with 86.0\% AUC. Regarding the gender recognition, see Fig. 4 (middle), good results are also achieved with 88.7\% AUC.

These preliminary results support the ability of machine learning systems to infer sensitive information of the subjects from the biometric raw data ($X$), which might be considered as an invasion of the personal privacy. The next experiments analyse the results achieved by the proposed GaitPrivacyON approach considering the privacy-preserving domain ($\widehat{X}$).

\subsection{GaitPrivacyON}

As indicated in Sec. 3.3, three different parameters can be configured in the training process of GaitPrivacyON to control the data transformation and the trade-off between the utility of the gait verification (main task) system and the sensitive information of the user (activity and gender): $\alpha$ (\textit{task loss} parameter), $\beta$ (\textit{content loss} parameter), and $\gamma$ (\textit{style loss} parameter).  

\subsubsection{MotionSense \& MobiAct Databases}

We first analyse the results achieved in the main task, mobile gait verification. Fig. \ref{fig:ROC_MotionMobi} (bottom) shows the ROC curves together with the AUC results of the gait biometrics verification system. Analysing the traditional approach, i.e., using the biometric raw data ($X$), the gait verification system is able to achieve accurate results with 99.9\% AUC over the final evaluation dataset. However, as it was commented in Sec. 5.1, from this traditional approach it is also possible to extract sensitive information, 99.2\% AUC for activity recognition and 99.0\% AUC for gender recognition.

The results achieved by GaitPrivacyON in the main task (gait verification) can be seen in Fig. \ref{fig:ROC_MotionMobi} (bottom). In general, we can see different AUC results depending on the values of the training parameters (including symbols), ranging from 99.6\% AUC to 91.0\% AUC. The selection of these parameters affects in the extraction of the activity and gender sensitive information.

Fig. \ref{fig:ROC_MotionMobi} (top) shows the ROC curves together with the AUC results achieved by GaitPrivacyON in the activity recognition task (dashes curves) when $X$ is replaced by $\widehat{X}$. It can be seen how the AUC results decrease as $\gamma$ increases, achieving a result close to random (49.02\% AUC) when $\gamma$ = 0.20.

A similar trend is also observed for the gender recognition task. Fig. \ref{fig:ROC_MotionMobi} (middle) shows the ROC curves together with the AUC results achieved by GaitPrivacyON in the gender recognition task (dashes curves). It can be seen how the AUC results decrease as $\gamma$ increases, achieving a result close to random (50.6\% AUC) when $\gamma$ = 0.30.

As a result, when the transformed data ($\widehat{X}$) provided by GaitPrivacyON achieves AUC values close to random (50.0\%) in the sensitive information tasks, it will be assumed to achieve privacy-preserving results, as long as the AUC of the gait verification task hardly decreases. Therefore, we select $\alpha= 0.40$, $\beta= 0.40$, $\gamma = 0.20$ as optimal configuration parameters, since the results on the gait biometrics verification task barely decrease (3.15\% AUC) while results close to random are achieved in both the activity (49.0\% AUC) and gender (57.2\% AUC).

\subsubsection{OU-ISIR Database}

Fig. \ref{fig:ROC_OUISIR} (bottom) shows the ROC curves together with the AUC results of the gait biometrics verification system. Using the biometric raw data ($X$) of the final evaluation dataset, the gait verification system is able to achieve accurate results with 97.5\% AUC. As in the previous case, with this traditional approach it is possible to extract much of the sensitive information such as the activity (86\% AUC) and gender (88.7\% AUC). For the OU-ISIR database, the best parameter configuration of GaitPrivacyON is $\alpha= 0.40$, $\beta= 0.50$, $\gamma = 0.10$. In this case, GaitPrivacyON achieves AUC results close to 50\% for both activity and gender recognition, while keeping a similar performance on gait verification compared with the traditional approach, i.e., 97.5\% AUC vs. 96.6\% AUC.

\subsection{Comparison with the State of the Art}

Analysing MotionSense and MobiAct databases together, GaitPrivacyON is able to decrease the AUC in the gender task (sensitive information) from 99.0\% to 57.2\% while reducing the performance from 99.9\% AUC to 96.7\% AUC in gait verification (main task). Moreover, using the OU-ISIR database, GaitPrivacyON also achieves robust results, decreasing the AUC from 88.7\% to 50.1\% in gender recognition while keeping similar AUC results in the main task, from 97.5\% to 96.6\%. In comparison to our work, the approach presented by Garofalo \textit{et al.} in \cite{garofalo2019data} using the OU-ISIR database decreased the F1-score in the gender recognition task from 73\% to 52\% while worsening the accuracy from 90.9\% to 85.3\% in the gait verification task. However, it is important to note that their method considers supervised learning, while GaitPrivacyON is based on unsupervised learning. 

Finally, for completeness, we highlight other approaches focused on the privacy-preserving of time sequences \cite{zhang2021preventing, boutet2021dysan, hajihassnai2021obscurenet}, although the topic is different, i.e., activity recognition. A similar trend can be observed when protecting sensitive information such as the age and identity of the person.

%\section{Limitations and Future Work}
%\label{sec:LimitationsandFutureWork}
%\input{LimitationsandFutureWork}

\section{Conclusions}
\label{sec:conclusions}

This study has presented GaitPrivacyON, a novel mobile gait biometrics verification approach that provides accurate authentication results while preserving the privacy of the subject. One of the main advantages of the approach is that the first module (convolutional Autoencoders) is trained in an unsupervised way, without specifying the sensitive attributes of the subject to protect. We have performed an in-depth quantitative analysis of GaitPrivacyON over three popular databases in the field of gait recognition, MotionSense \cite{malekzadeh2018protecting}, MobiAct \cite{vavoulas2016mobiact}, and OU-ISIR \cite{Ngo2019OUISIR}. Our model is able to obtain good results, as the gait biometrics verification task barely decrease (3.2\% AUC with MotionSense and MobiAct databases and 0.9\% with OU-ISIR database) while results close to random are achieved in both the activity and gender ($\sim$50\% AUC) tasks. In conclusion, GaitPrivacyON increases the protection of the sensitive data (e.g., activity and gender) with unsupervised learning while being able to maintain the accuracy of the gait biometrics verification task. The proposed GaitPrivacyON approach have been evaluated with discrete sensitive attributes (i.e., activity and gender) and further experiments are necessary to adapt the method to continuous sensitive attributes (e.g., weight or age). Our approach is based on a semi-supervised learning approach and therefore, it requires large amount of labelled data (sensitive attributes). Future work will be oriented to: 1) reduce the amount of data needed to train the models using unsupervised approaches; 2) reduce the training time through GPU parallelization.

%GaitPrivacyON is a challenge to sensitive data inference attacks in the world of gait biometrics verification, as it uses unsupervised learning to remove various kinds of sensitive information from both the raw data and the model, making this information practically random while preserving the gait biometrics verification task.

\section{Acknowledgements}
This project has received funding from the European Union’s Horizon 2020 research and innovation programme under the Marie Skłodowska-Curie grant agreement No 860315. R. Tolosana and R. Vera-Rodriguez are also supported by INTER-ACTION (PID2021-126521OB-I00 MICINN/FEDER).
\bibliography{mybibfile}

\end{document}